\documentclass[runningheads]{llncs}
\usepackage[T1]{fontenc}
\usepackage{graphicx,verbatim}
\usepackage{multirow}
\usepackage{cite}
\usepackage{booktabs}
\usepackage{subcaption}
\usepackage{array}
\usepackage{float}
\usepackage{adjustbox}
\usepackage{amsmath}
\usepackage{threeparttable}
\usepackage{amssymb}
\usepackage{wrapfig}
\usepackage{enumitem}
\usepackage{svg}
\usepackage{marvosym}
\usepackage{placeins}
\usepackage{caption}
\usepackage{makeidx}
\usepackage{hyperref}
\hypersetup{
    colorlinks=true,    % 激活链接的颜色
    linkcolor=blue,     % 设置链接的颜色
    citecolor=blue,     % 设置引用的颜色
    filecolor=blue,     % 设置文件链接的颜色
    urlcolor=blue,      % 设置 URL 链接的颜色
    pdfborder={0 0 0}     % 保留颜色，但不显示链接框
}

\newcommand{\LetterMark}{\textsuperscript{(\Letter)}}

\begin{document}
\title{DiffOSeg: Omni Medical Image Segmentation via Multi-Expert Collaboration Diffusion Model}
\titlerunning{DiffOSeg: Omni Medical Image Segmentation}
% \author{
% Han Zhang\inst{1} %index{Zhang, Han}
% \and 
% Xiangde Luo\inst{1}
% \and 
% Yong Chen\inst{1} %index{Chen, Yong} 
% \and 
% Kang Li\inst{1,2}\textsuperscript{(\Letter)} %index{Li, Kang}
% }
\author{
Han Zhang\inst{1} \and 
Xiangde Luo\inst{1} \and 
Yong Chen\inst{1} \and 
Kang Li\inst{1}\inst{2}\LetterMark
}

\authorrunning{Zhang et al.}
\institute{West China Biomedical Big Data Center, West China Hospital, Sichuan University 
\and
Shanghai Artificial Intelligence Laboratory, Shanghai, China% 
\\ \email{likang@wchscu.cn}}%
\maketitle              % typeset the header of the contribution
\begin{abstract}
Annotation variability remains a substantial challenge in medical image segmentation, stemming from ambiguous imaging boundaries and diverse clinical expertise. Traditional deep learning methods producing single deterministic segmentation predictions often fail to capture these annotator biases. Although recent studies have explored multi-rater segmentation, existing methods typically focus on a single perspective\textemdash either generating a probabilistic ``gold standard'' consensus or preserving expert-specific preferences\textemdash thus struggling to provide a more omni view.
In this study, we propose DiffOSeg, a two-stage diffusion-based framework, which aims to simultaneously achieve both consensus-driven (combining all experts' opinions) and preference-driven (reflecting experts' individual assessments) segmentation.
Stage I establishes population consensus through a probabilistic consensus strategy, while Stage II captures expert-specific preference via adaptive prompts. 
Demonstrated on two public datasets (LIDC-IDRI and NPC-170), our model outperforms existing state-of-the-art methods across all evaluated metrics. Source code is available at \url{https://github.com/string-ellipses/DiffOSeg}.
\end{abstract}
\section{Introduction}\label{sec:Introduction}
In medical image segmentation, ambiguous target boundaries or irregular shapes often lead to inter-expert annotation variability~\cite{ref_cmr, ref_lesion_seg, ref_oncologist}. Essentially, this variability arises from two main factors: inherent image ambiguities and expert-specific preferences. 
For example, tumor boundaries are often ill-defined due to grayscale gradients or imaging artifacts, creating ambiguous annotation criteria. While subjectivity persists, multi-expert consensus can help define "relatively reasonable" boundaries. Besides, experts' annotation styles exhibit distinct variations--radiation oncologists may prioritize broader margins for tumor coverage~\cite{ref_oncologist}, while surgeons prefer precise boundaries for tissue protection~\cite{ref_surgeon}. Therefore, it is essential to simultaneously address the dual demands for capturing consensus and preserving expertise.

Traditional segmentation networks~\cite{ref_unet, ref_medsam} typically generate single deterministic masks, often inadequate for clinical needs. Recent studies~\cite{ref_staple, ref_cmr, ref_prob_unet, ref_ccdm2, ref_phiseg, ref_cm_pixels} have explored multi-rater segmentation, which can be generally categorized into three paradigms: meta-segmentation, diversified segmentation, and personalized segmentation. Meta-segmentation~\cite{ref_staple} reconstructs a pseudo gold standard through annotation fusion, though the existence of such a gold standard still remains debated in many medical contexts~\cite{ref_cmr, ref_lesion_seg}. Diversified segmentation~\cite{ref_prob_unet, ref_ccdm2, ref_phiseg} learns the distribution of multi-expert consensus and then samples multiple plausible annotations from it. Personalized segmentation~\cite{ref_pionono, ref_cm_pixels} captures expert-specific styles, producing annotations tailored to individual diagnostic preferences. While both meta- and diversified segmentation prioritize capturing consensus-driven patterns, they fail to preserve expert individuality. Conversely, preference-driven approaches (personalized segmentation) focus on individual experts’ styles while ignoring shared clinical insights. 
% 提一下TAB和PADL
Recent studies~\cite{liao2023transformer, liao2024modeling} have attempted to unify consensus and personalized segmentation, but their deterministic point estimation overlooks expert annotation diversity. Although D-Persona~\cite{ref_DPersona} attempts to address this limitation, its performance is constrained by a limited expressive latent space and redundant expert-specific projection heads, which overlook inter-expert correlations.
% Consequently, existing methods thus lack an omni perspective to unify probabilistic consensus and preference-driven segmentation. 

%增加DPMs vs VAE?
Inspired by the success of Diffusion Probabilistic Models (DPMs)~\cite{ref_ddpm, ref_ddim} in capturing complex data distribution and excelling in conditional generation tasks, we propose DiffOSeg, a diffusion-based multi expert collaboration framework that unifies population-level consensus learning and expert-specific preference adaptation
Built on a categorical diffusion model\cite{ref_ccdm2, ref_ccdm1, ref_tabddpm}, our approach operates in two learning stages: Stage I applys a probabilistic consensus strategy to integrate multi-expert consensus, enhancing segmentation diversity and generalization. 
Stage II employs learnable preference prompts that encode expert-specific styles through a plug-in prompt block, steering denoising toward expert-specific segmentation patterns. The framework achieves parameter efficiency with 13.1 M parameters, representing a 54\% reduction compared to D-Persona's 28.46 M. 
Extensive experiments on the LIDC-IDRI and NPC-170 datasets demonstrate DiffOSeg's effectiveness in both consensus-driven and preference-driven segmentation tasks, achieving state-of-the-art performance across all evaluated metrics.

Our contributions includes: (1) First diffusion-based framework unifying consensus and preference-driven segmentation; (2) Stage I: Probabilistic consensus strategy integrating multi-rater agreements while preserving distributional diversity; (3) Stage II: Adaptive preference prompting mechanism dynamically encoding expert-specific annotation patterns; (4) Superior performance demonstrated on LIDC-IDRI and NPC-170 datasets for both segmentation paradigms.

\section{Method}
% \subsection{Overview} 
\begin{figure}[t]
    \centering
    % 使用graphicx包对PDF图像裁剪
    \includegraphics[width=\linewidth, trim=0cm 5.4cm 10.2cm 0cm, clip]{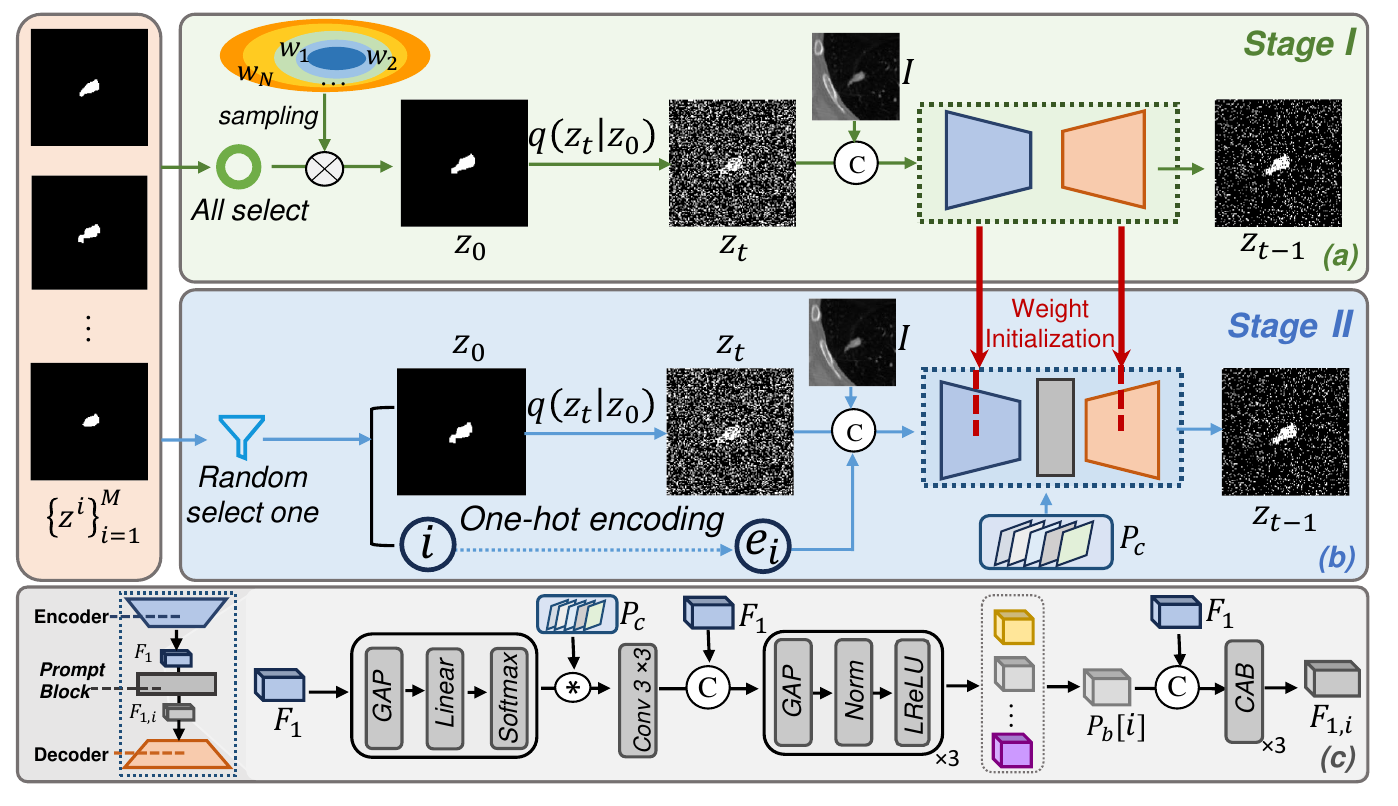}
    \caption{\textbf{DiffOSeg pipeline.} 
    (a) Stage I (Training): Probabilistic consensus stragety dynamiclly integrates multi-expert annotations (Sec~\ref{sec:consensus_segmentation}); (b) Stage II (Training): Adaptive prompts model expert preferences (Sec~\ref{sec:preference_segmentation}); (c) Plug-in prompt block (Sec~\ref{sec:preference_segmentation}).}
    % (b) Stage II (Training): Preference prompts are integrated into the model through a plug-in prompt block to enable the generation of personalized segmentation masks (Sec~\ref{sec:personalized_segmentation}); 
    % % 从所有不同专家的标注中中随机抽取一个，将对应专家的身份进行编码。将被破坏的标注
    % % (c) The encoder feature $F_1$ is used to weight the prompt parameters. These weighted prompts are concatenated with $F_1$ and processed through three GAP-Norm-LReLU layers to generate dynamic prompts. The relevant prompt for the target annotator is selected, concatenated with $F_1$, and passed into the decoder.  
    % (c) Stage I\&II (Inference). Both stages start with randomly sampled noise and perform $T$ denoising steps to produce a clean mask. The difference lies in the condition and the denoising UNet (Sec~\ref{sec:personalized_segmentation}). Please zoom in for details.}  
    \label{fig:main_workflow}
\end{figure}
% 和Introduction相比，多介绍一些思想和high-level层面的东西
%% 强调框架设计哲学：共识与偏好的协同建模
% The proposed framework provides an omni-solution for medical image segmentation through a unified diffusion-based architecture
% The proposed framework is designed to address the dual demands of medical image segmentation through a unified diffusion-based architecture: capturing population-level consensus while preserving individual preferences. As illustrated in Figure \ref{fig:main_workflow}, our proposed DiffOSeg is built on a conditional categorical diffusion model \cite{ref_ccdm1, ref_ccdm2} and operates through two stages, which address consensus integration and personalized adaptation respectively.
\subsection{Consensus-Driven Segmentation}\label{sec:consensus_segmentation}
% 强调概率共识策略与群体共识建模的关系
When annotating the same image, different experts exhibit distinct annotation patterns, reflecting both shared insights and individual preferences. Here raises a key question: How can we effectively capture inter-expert consensus while preserving diversity? To address this, we propose a probabilistic consensus strategy that dynamically integrates multi-expert annotations, enabling the model to reconcile both aspects by considering different expert subsets.

Given an image $I \in \mathbb{R}^{C \times H \times W}$ with $M$ expert annotations ${\{\mathbf{z}^{i}\}}_{i=1}^{M}$, each annotation $\mathbf{z}^{i} \in  \{e_1, \dots, e_L\}^{K}  $ is structured as a $K 
\times L$ matrix. Here, each row corresponds to a one-hot encoded label, $K$ denotes the number of pixels, and $\mathcal{L} = \{1, \dots, L\}$ represents the set of discrete labels. For example, if the $k$-th pixel is labeled $l$, then
\begin{equation}
\mathbf{z}^{i}[k, :] = e_l \in \{0, 1\}^{L}, 
\end{equation}
Here $e_{l}$  with 1 in position $l$ and 0 elsewhere.
To balance the contributions of multiple experts, we define a probabilistic weight distribution $W$, which generates a random vector $\mathbf{w} \in \left[0, 1\right ]^{M}$ that determines the importance of the $M$ experts. The support set $\mathcal{S}$ of $W$ includes all possible configurations of $\mathbf{w}$, representing the power set of the annotation space:
\begin{equation}
\mathcal{S} = \{ \mathbf{w} \mid \mathbf{w} \in \left[0, 1\right ]^M, \|\mathbf{w}\|_1 \leq M \}.
\end{equation}
Here, $|\mathbf{w}|_1$ represents the number of active experts contributing to the consensus for a given sample. By sampling $\mathbf{w}$ from $S$, the model dynamically adjusts the participation of experts, enabling three distinct scenarios: 1) \textbf{Single Vote}: Only one expert contributes to the consensus, represented by $\|\mathbf{w}\|_1 = 1$; 2) \textbf{Subgroup Consensus}: A subset of experts contributes, where $\|\mathbf{w}\|_1 = k$ and $1 < k < M$; 3) \textbf{Full Vote}:  All experts participate equally, represented by $\|\mathbf{w}\|_1 = M$. 

The consensus label $\mathbf{z}^{c}$ is computed as a weighted linear combination of expert annotations $\left [\mathbf{z}^{1}, \mathbf{z}^{2}, \dots, \mathbf{z}^{M} \right ]$, scaled by the probabilistic weight vector $\mathbf{w}$, as formulated in Eq.~\eqref{mixed_label} (left).
%每个元素都有均等的概率为0或1. %图
We treated all latent variables $\mathbf{z}^{c}_{0:T}$ as categorical and modeled both forward and reverse transition distributions using categorical distribution. The initial state $\mathbf{z}^{c}_{0}$ is set to the consensus label $\mathbf{z}^{c}$. The reverse (generative) process is expressed as Eq.~\eqref{mixed_label} (right).
% \begin{align}
%  \mathbf{z}^{c} &= \mathbf{w} \times \left [\mathbf{z}^{1}, \mathbf{z}^{2}, \dots, \mathbf{z}^{M} \right ]^{T},  &
%  p_{\theta }\left ( \mathbf{z}^{c}_{0:T} \mid I\right ) &= p\left ( \mathbf{z}^{c}_T\right )\displaystyle\prod_{t=1}^{T}p_{\theta}\left(\mathbf{z}^{c}_{t-1} \mid \mathbf{z}^{c}_{t}, I\right),
% \end{align}
\begin{equation}\label{mixed_label}
  \begin{aligned}
    \mathbf{z}^{c} &= \mathbf{w} \times \left[ \mathbf{z}^{1}, \mathbf{z}^{2}, \dots, \mathbf{z}^{M} \right]^{T}, &
    p_{\theta}\left( \mathbf{z}^{c}_{0:T} \mid I \right) &= p\left( \mathbf{z}^{c}_T \right) \prod_{t=1}^{T} p_{\theta}\left( \mathbf{z}^{c}_{t-1} \mid \mathbf{z}^{c}_{t}, I \right),
  \end{aligned}
\end{equation}
where the reverse transition probability is defined as (for clarity, we omit the superscript $c$ in $\mathbf{z}^{c}_{0:T}$):
\begin{equation}\label{derivations}
    \begin{aligned}
    % \begin{equation} \label{eq:reverse_transition}
    p_{\theta}\left(\mathbf{z}_{t-1} \mid \mathbf{z}_{t}, I\right) &=
    \begin{cases}
    \mathcal{C}(\mathbf{z}_{0}; \hat{\mathbf{p}}_0) & \text{if } t=1, \\
    \sum_{\mathbf{z}_0} \mathcal{C}(\mathbf{z}_{t-1}; \pi(\mathbf{z}_t, \mathbf{z}_0)) \cdot \mathcal{C}(\mathbf{z}_0; \hat{\mathbf{p}}_0) & \text{otherwise}.
    \end{cases} \\
    \pi(\mathbf{z}_t, \mathbf{z}_0) &= \frac{1}{\tilde{\pi}} \left( \frac{\beta_t}{L} \mathbf{1} + \alpha_t \mathbf{z}_t \right) \odot \left( \frac{1 - \bar{\alpha}_{t-1}}{L} \mathbf{1} + \bar{\alpha}_{t-1} \mathbf{z}_0 \right),
    % \end{equation}
    \end{aligned}
\end{equation}
where
$\tilde{\pi} = \frac{1 - \bar{\alpha}_t}{L} + \bar{\alpha}_t \cdot \delta_{z_t}^{z_0}$, $\delta$ is the Kronecker delta and $\odot$ denotes element-wise multiplication.
For detailed derivations about Eq.~\eqref{derivations}, please refer to \cite{ref_ccdm1, ref_ccdm2}.

To approximate $p_{\theta}\left(\mathbf{z}_{t-1} \mid \mathbf{z}_{t}, I\right)$, we use a neural network to predict $\hat{\mathbf{p}}_0$. And the training objective is to minimize the Kullback-Leibler (KL) divergence between the forward process posterior $q(\mathbf{z}_{t-1} \mid \mathbf{z}_t, \mathbf{z}_0)$ and the predicted reverse distribution $p_{\theta}(\mathbf{z}_{t-1} \mid \mathbf{z}_t, I)$, ensuring the model effectively approximates the true reverse process:
\begin{equation}
    \begin{aligned}
     \hat{\mathbf{p}}_0 &= f_{\theta}(\mathbf{z}_t, t, I) \in [0, 1]^{K \times L}, \\
     KL(q \parallel p) &= KL\left( q(\mathbf{z}_{t-1} \mid \mathbf{z}_t, \mathbf{z}_0) \parallel p_\theta(\mathbf{z}_{t-1} \mid \mathbf{z}_t, I) \right).
    \end{aligned}
\end{equation}
During inference, the denoising process begins with randomly sampled Gaussian noise and iteratively refines over $T$ steps, guided by the corresponding image condition through the backbone denoising UNet (``backbone'' is used to distinct from Stage II’s one). 
% 4.浅析影响
% 补充具体实例或可视化
% 通过举例说明 $W$ 的作用，例如：
% 有三个标注者对某像素的标注分别是 ${1, 2, 1}$，$W$ 的不同取值（如 $[0.2, 0.3, 0.5]$ 或 $[0.33, 0.33, 0.33]$）会如何影响最终的 $\mathbf{z}_0$。
% 使用图示展示 $W$ 对共识生成的影响，比如标注分布的可视化。
% 总结这种策略的实际意义
% 强调它不仅提高了一致性（consistency），还保留了差异性（diversity），从而增强了模型的泛化能力。
% 指出这种策略如何帮助模型更好地适应多标注者环境下的不确定性。
% 注意两个阶段的递进
% 连贯性
% \begin{textblock}{10}(0.75,0.05)  % 设置图像位置，(x,y) 坐标，x 为页面宽度的 75%，y 为页面高度的 5%
%     \includegraphics[clip, trim=620pt 60pt 0pt 0pt,width=0.25\textwidth]{svg_image/pip.pdf}  % 设置图像宽度
% \end{textblock}
\subsection{Preference-Driven Segmentation} 
\label{sec:preference_segmentation}
% 建议个性化分割与共识建模的协同关系
% \begin{wrapfigure}{r}{0.5\linewidth} % 'r' 表示图像放右侧
%     \centering
%     \includegraphics[clip, trim=630pt 55pt 0pt 0pt, width=\linewidth]{svg_image/pip.pdf}  % 设置图像宽度
%     \caption{The architecture of our proposed plug-in prompt block.}  % 图像标题
%     \label{fig:pip}
% \end{wrapfigure}
% 在第二阶段中，我们的目标是希望模型可以捕捉到各个标注者的偏好。 我们设计了一个身份模块，用于编码每个评估者的身份信息，并将其注入到网络当中。 
% 凸显diffusion model特性
% 凸显prompt techniques特性
Building upon Stage I's consensus features, Stage II specializes in capturing expert-specific deviations that reflect their unique annotation styles. We employ prompting-based techniques~\cite{ref_vpt, ref_promptir}, where tunable prompts encode individual preference of each expert. These prompts act as dynamic adapters, allowing the model to modulate its behavior based on the target expert's identity, without requiring separate networks for each expert. 

As shown in Fig.~\ref{fig:main_workflow}(b), during each training step, we randomly sample an expert annotation $z_i$ per image and encode its identity $i$ as a one-hot vector, as described in \cite{ref_fip, ref_dodnet}. This vector and the corresponding image are initially concatenated with $z_t$. Then, to enhance feature specificity, our plug-in prompt block (Fig.~\ref{fig:main_workflow}(c)) injects expert-specific prompts into features from the encoder, enabling dynamic adaptation to individual annotation styles. 
The prompt integration mechanism operates through two sequential phases: 
\subsubsection{Prompt-Weight Generation.}To generate context-aware prompt-weights, we first extract global context from input features $F_{1}$ via spatial average pooling (GAP). This pooled feature is projected to a lower-dimensional space through a linear layer, followed by $softmax$ normalization to produce fusion weights $\lambda \in \mathbb{R}^{N}$. These weights dynamically adjust the prompt components $P_c$ through a $3 \times 3$ convolution. The process is summarized as:
\begin{equation}
\begin{small}
\begin{aligned} 
    \lambda &= \text{Softmax}\left( \text{Linear}\left( \text{GAP}\left( F_{1}\right )\right ) \right), & P_m &= \text{Conv}_{3\times 3}(\lambda P_{c}) 
\end{aligned}
\end{small}
\end{equation}
\subsubsection{Feature Modulation.}The modulated prompts $P_m$ are concatenated with $F_1$ and passed through three GAP-Norm-LReLU blocks to generate dynamic expert-specific weights $P_b$. For the $i$-th expert, the corresponding prompt $P_b[i]$ is selected and integrated with backbone features $F_1$ via cascaded Channel Attention Blocks (CABs)\cite{ref_promptmr}, which results in the final prompted feature $F_{1,i}$:
\begin{equation}
\begin{small}
\begin{aligned} 
    P_{b} &= \text{Enhance}(P_m, F_1), 
    & F_{1,i} &= \text{Modulate}(P_b[i], F_1). 
\end{aligned}
\end{small}
\end{equation}

The training objective remains the same as in Sec.~\ref{sec:consensus_segmentation}. During inference, the process starts from randomly sampled noise, with the image and one-hot encoded identity vector serving as the condition to guide the prompt-driven denoising UNet. After $T$ denoising steps, the model outputs a complete segmentation mask that reflects the target expert's annotation style. 
% By leveraging the iterative refinement properties of diffusion models, preference prompts enable the network to dynamically adapt to individual experts' preferences, achieving a balance between personalization and generalization.
\section{Experiments}
% 添加对LIDC-IDRI的解释
\subsection{Experimental Setup}
\subsubsection{Dataset.} 
We evaluate our method on two benchmark datasets: LIDC-IDRI and NPC-170. The \textbf{LIDC-IDRI} dataset\cite{ref_lidc_idri} consists of 1,609 2D thoracic CT scans from 214 subjects with manual annotations from four domain experts. To simulate the expert preferences, we manually rank the four annotated regions following the experimental protocols established in \cite{ref_cm_pixels, ref_DPersona}. Experiments are conducted using the 4-fold cross-validation at the patient level, as described in \cite{ref_medicalmatting, ref_DPersona}. The \textbf{NPC-170} dataset\cite{ref_npc170, ref_mmis2024} includes 170 Nasopharyngeal Carcinoma (NPC) MRI subjects with GTVp annotations from 4 senior radiologists. It is split into training (100 subjects), validation (20 subjects), and testing (50 subjects) sets, following the setup in \cite{ref_mmis2024, ref_DPersona}.
\subsubsection{Implementation Details.} 
All experiments are conducted on a single NVIDIA GeForce RTX 4090 GPU, with total 13.1 M parameters \textemdash significantly 54\% fewer than D-Persona's 28.46 M\cite{ref_DPersona}. Stage II is initialized using Stage I’s pretrained weights and fine-tuned end-to-end with all parameters trainable.
Both datasets are resized to $128 \times 128$ resolution with [0,1] intensity normalization. The diffusion process employs $T$ = 250 steps under cosine noise scheduling.  
In the training phase, we use the Adam optimizer (lr=\(10^{-4}\)) with batch size 12. To enhance generalization, data augmentation techniques such as random rotations, flips, and intensity scaling are applied. For LIDC-IDRI, the model is trained for 120,000 iterations in Stage I and 9,0000 iterations in Stage II. For NPC-170, both Stage I and Stage II are trained for 400,000 iterations.
%% 方法
%% 评价指标
\subsection{Consensus-Driven Segmentation}
\begin{figure}[t]
\centering
\noindent
\begin{minipage}{\linewidth}
    \centering
    % Table 放在这里
    \captionof{table}{Consensus-driven segmentation performance comparison ($Dice_{soft}$ are denoted as $D^{s}_{n}$ and shown as percentages). Results for DiffOSeg and PhiSeg are obtained from our implementation, and others are cited from\cite{ref_DPersona}. Best in \textbf{bold}.}
    \label{tab:diversity_performance}
    \resizebox{0.95\linewidth}{!}{
    \begin{tabular}{ccllllcllllll}
    \cline{1-9}
    \multicolumn{1}{c|}{\multirow{2}{*}{\textbf{Method}}} & \multicolumn{6}{c}{\textbf{LIDC-IDRI}}                                                                                                                & \multicolumn{2}{c}{\textbf{NPC-170}}                     \\  
    \multicolumn{1}{c|}{}                                 &  $GED_{\scriptscriptstyle 10}(\downarrow)$               & \multicolumn{1}{c}{$GED_{\scriptscriptstyle 30}(\downarrow)$}            & \multicolumn{1}{c}{$GED_{\scriptscriptstyle 50}(\downarrow)$}            & \multicolumn{1}{c}{$D^{s}_{\scriptscriptstyle 10}(\uparrow)$}            & \multicolumn{1}{c}{$D^{s}_{\scriptscriptstyle 30}(\uparrow)$}            & \multicolumn{1}{c|}{$D^{s}_{\scriptscriptstyle 50}(\uparrow)$}            & \multicolumn{1}{c}{$GED_{\scriptscriptstyle 30}(\downarrow)$}           & \multicolumn{1}{c}{$D^{s}_{\scriptscriptstyle 30}(\uparrow)$}   \\ \cline{1-9}
    \multicolumn{1}{c|}{Prob. UNet\cite{ref_prob_unet}}                       & 0.2181              & \multicolumn{1}{c}{0.2169}           & \multicolumn{1}{c}{0.2168}           & \multicolumn{1}{c}{88.79}          & \multicolumn{1}{c}{88.79}          & \multicolumn{1}{c|}{88.80}          & \multicolumn{1}{c}{0.4466}          & \multicolumn{1}{c}{75.34} \\
     \multicolumn{1}{c|}{PhiSeg\cite{ref_phiseg}}              & 0.1934             & \multicolumn{1}{c}{0.1921}           & \multicolumn{1}{c}{0.1917}           & \multicolumn{1}{c}{89.31}          & \multicolumn{1}{c}{89.73}          & \multicolumn{1}{c|}{89.92}          & \multicolumn{1}{c}{-}          & \multicolumn{1}{c}{-} \\
    \multicolumn{1}{c|}{D-Persona (Stage I)\cite{ref_DPersona}}              & 0.1461              & \multicolumn{1}{c}{0.1375}           & \multicolumn{1}{c}{0.1358}           & \multicolumn{1}{c}{90.24}          & \multicolumn{1}{c}{90.42}          & \multicolumn{1}{c|}{90.45}          & \multicolumn{1}{c}{0.2385}          & \multicolumn{1}{c}{\textbf{80.40}} \\
    \hline
    \multicolumn{1}{c|}{DiffOSeg (Stage I)}                   & \textbf{0.09154}    & \multicolumn{1}{c}{\textbf{0.0773}} & \multicolumn{1}{c}{\textbf{0.0745}} & \multicolumn{1}{c}{\textbf{91.21}} & \multicolumn{1}{c}{\textbf{92.20}} & \multicolumn{1}{c|}{\textbf{92.37}} & \multicolumn{1}{c}{\textbf{0.1822}} & \multicolumn{1}{c}{79.83} \\ \cline{1-9}
    \end{tabular}}
\end{minipage}

\begin{minipage}{\linewidth}
    \centering
    % Figure 放在这里
    \includegraphics[width=0.85\linewidth, trim=0 165 0 0, clip]{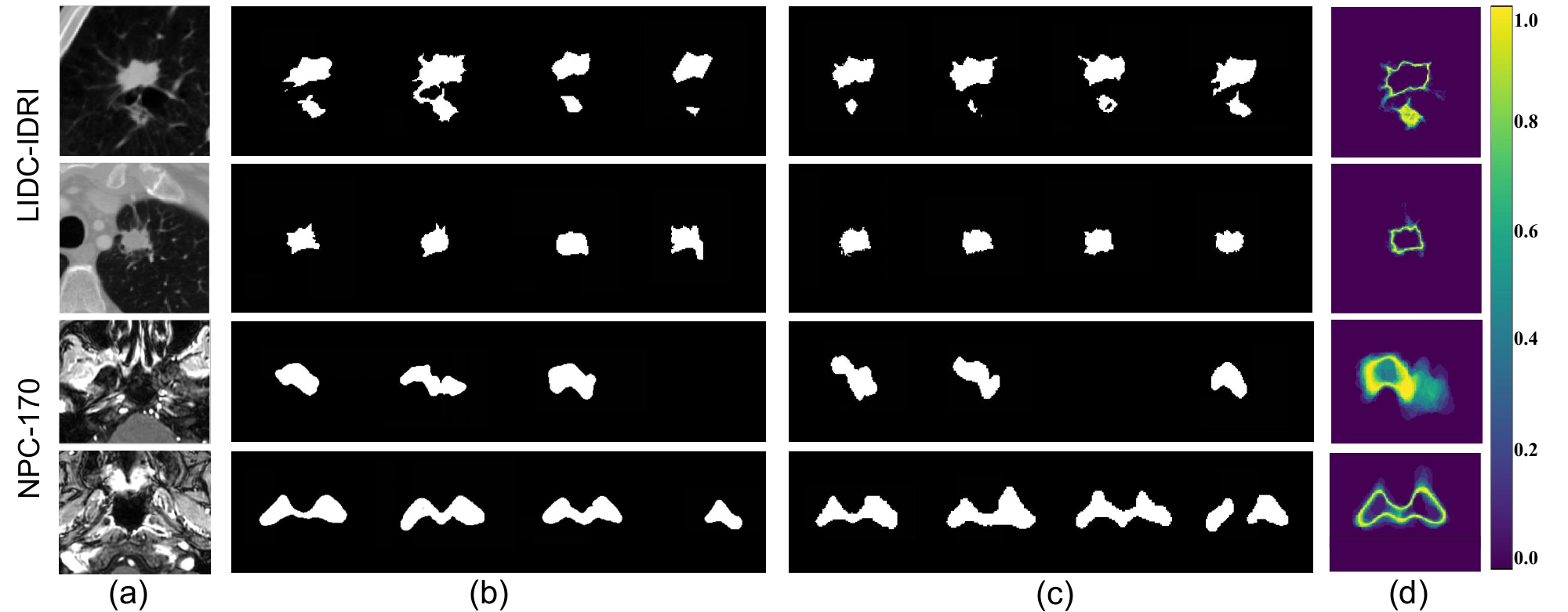}
    \caption{Consensus-driven segmentation visualization on both datasets.
    (a) Original images, (b) expert annotations, (c) sampled DiffOSeg predictions for multi-expert consensus, (d) corresponding uncertainty maps (brighter = higher unncertainty).}
    \label{fig:diversified_segmentation}
\end{minipage}
\end{figure}
 % The average trend is emphasized rather than a rare peak, as low-probability events are considered unreliable from a statistical perspective. 
% \subsubsection{Baselines and Metrics.}
% 明确说明是和概率性分割方法比较
% 对GED和Die_soft的阐释更明确
In Stage I, we compare DiffOSeg with some representative probabilistic segmentation methods, including Prob. UNet~\cite{ref_prob_unet}, PhiSeg~\cite{ref_phiseg} and D-Persona~\cite{ref_DPersona}.
We employ two complementary metrics: the Generalized Energy Distance ($GED$) to assess segmentation diversity, and Threshold-Aware Dice ($Dice_{soft}$)~\cite{ref_DPersona} to evaluate segmentation fidelity. For $Dice_{soft}$, the thresholds are set as $\{0.1, 0.3, 0.5, 0.7, 0.9\}$.
% $GED$ quantifies the alignment between the predicted distribution and the expert annotations' distribution, as well as the variability within the predicted distribution itself. $Dice_{soft}$ computes $Dice$ coefficient across multiple thresholds, providing a more effective reflection of the model's performance in handling ambiguous or uncertain regions. 
We denote the metrics computed with $n$ samples using a subscript, i.e., $GED_n$ and $D^{s}_n$, and $n$ is set to common values found in the literature ~\cite{ref_ccdm2, ref_DPersona, ref_mose}. 
% 添加对GED等指标的说明

As shown in Table~\ref{tab:diversity_performance}, on LIDC-IDRI (4-fold cross-validation average), DiffOSeg achieves superior improvements in $GED$ metrics across all sample sizes (10/30/50), outperforming the second-best method by 37.4\% for $GED_{10}$ (0.0915 vs. 0.1461), 43.8\% for $GED_{30}$ (0.0773 vs. 0.1375), and 45.2\% for $GED_{50}$ (0.0745 vs. 0.1358). 
For $Dice_{soft}$, it also exhibits great performance across all sample sizes. On NPC-170, DiffOSeg reduces $GED_{30}$ by 23.6\% compared to D-Persona (0.1822 vs. 0.2385), though it slightly underperforms D-Persona on $D^{s}_{30}$.
Visualization results shown in Fig.~\ref{fig:diversified_segmentation} demonstrate our model's ability to capture expert consensus while appropriately preserving diversity.

\subsection{Preference-Driven Segmentation}
% \begin{figure}[t]
% \includegraphics[width=0.9\linewidth, trim=0 125 0 0, clip]{svg_image/2.pdf}
% \caption{Personalized segmentation results on LIDC-IDRI and NPC-170.} 
% \label{fig:personalized_segmentation}
% \end{figure}
\begin{figure}[t]
\centering
\noindent
% Table部分
\begin{minipage}{\linewidth}
    \centering
    \renewcommand{\arraystretch}{1.1}
    \captionof{table}{Preference-driven segmentation performance comparison (all metrics shown as percentages). Results for DiffOSeg and PhiSeg are obtained from our implementation, and other scores are cited from\cite{ref_DPersona}. Best in \textbf{bold}.}
    \label{tab:personality_performance}
    \resizebox{\textwidth}{!}{ % 调整宽度为页面宽度
    \begin{tabular}{ccccccccccc}
    \toprule
    \multicolumn{1}{c|}{\multirow{2}{*}{\textbf{Method}}} & \multicolumn{5}{c}{\textbf{LIDC-IDRI}}                                                                                    & \multicolumn{5}{c}{\textbf{NPC-170}}                                                                             \\
    \multicolumn{1}{c|}{}                                 & $D_{A_{1}} (\uparrow)$               & $D_{A_{2}} (\uparrow)$               & $D_{A_{3}} (\uparrow)$               & $D_{A_{4}} (\uparrow)$               & \multicolumn{1}{c|}{$D_{mean} (\uparrow)$} & $D_{A_{1}} (\uparrow)$               & $D_{A_{2}} (\uparrow)$               &$D_{A_{3}} (\uparrow)$               & $D_{A_{4}} (\uparrow)$                & $D_{mean} (\uparrow)$            \\ \midrule

    \multicolumn{1}{c|}{CM-Global\cite{ref_cmglobal}}                        &     86.13               &     88.76                 &      88.99                &         86.18             & \multicolumn{1}{c|}{87.51}         &       77.92               &      74.65                &        71.97              &        75.47              &        75.00   \\
    \multicolumn{1}{c|}{CM-Pixel\cite{ref_cm_pixels}}                         &     85.99                 &      88.81                &          89.31            &        86.77              & \multicolumn{1}{c|}{87.72}          &       78.97               &         73.93             &      71.65                &        75.12              &        74.92              \\
    \multicolumn{1}{c|}{TAB\cite{ref_TAB}}                              &        85.00              &     86.35                 &        86.77              &       85.77               & \multicolumn{1}{c|}{85.97}          &       77.33               &           73.45           &      74.83                &        71.67              &       74.32              \\
    \multicolumn{1}{c|}{Pionono\cite{ref_pionono}}                          &       87.94               &   89.11                   &         89.55             &       88.76               & \multicolumn{1}{c|}{88.84}         &         78.87             &         74.11             &         71.97             &          75.41            &       75.09                \\
    \multicolumn{1}{c|}{D-Persona (Stage II)\cite{ref_DPersona}}             &        88.54              &     89.50                &      90.03                &         88.60             & \multicolumn{1}{c|}{89.17}         &         79.78             &         \textbf{74.60}             &         75.22             &          \textbf{75.17}            &       76.19               \\ \midrule
    \multicolumn{1}{c|}{DiffOSeg (Stage II)}                             &     \textbf{91.85}                 &       \textbf{90.15}               &          \textbf{91.3}            &       \textbf{90.68}               & \multicolumn{1}{c|}{\textbf{90.99}}         &        \textbf{81.16}              &        74.03              &          \textbf{76.47}            &       75.07               &         \textbf{76.88}             \\ 
    \hline
    \end{tabular}
    }
\end{minipage}

% Figure部分
\begin{minipage}{\linewidth}
    \centering
    \includegraphics[width=0.9\linewidth, trim=0 118 0 0, clip]{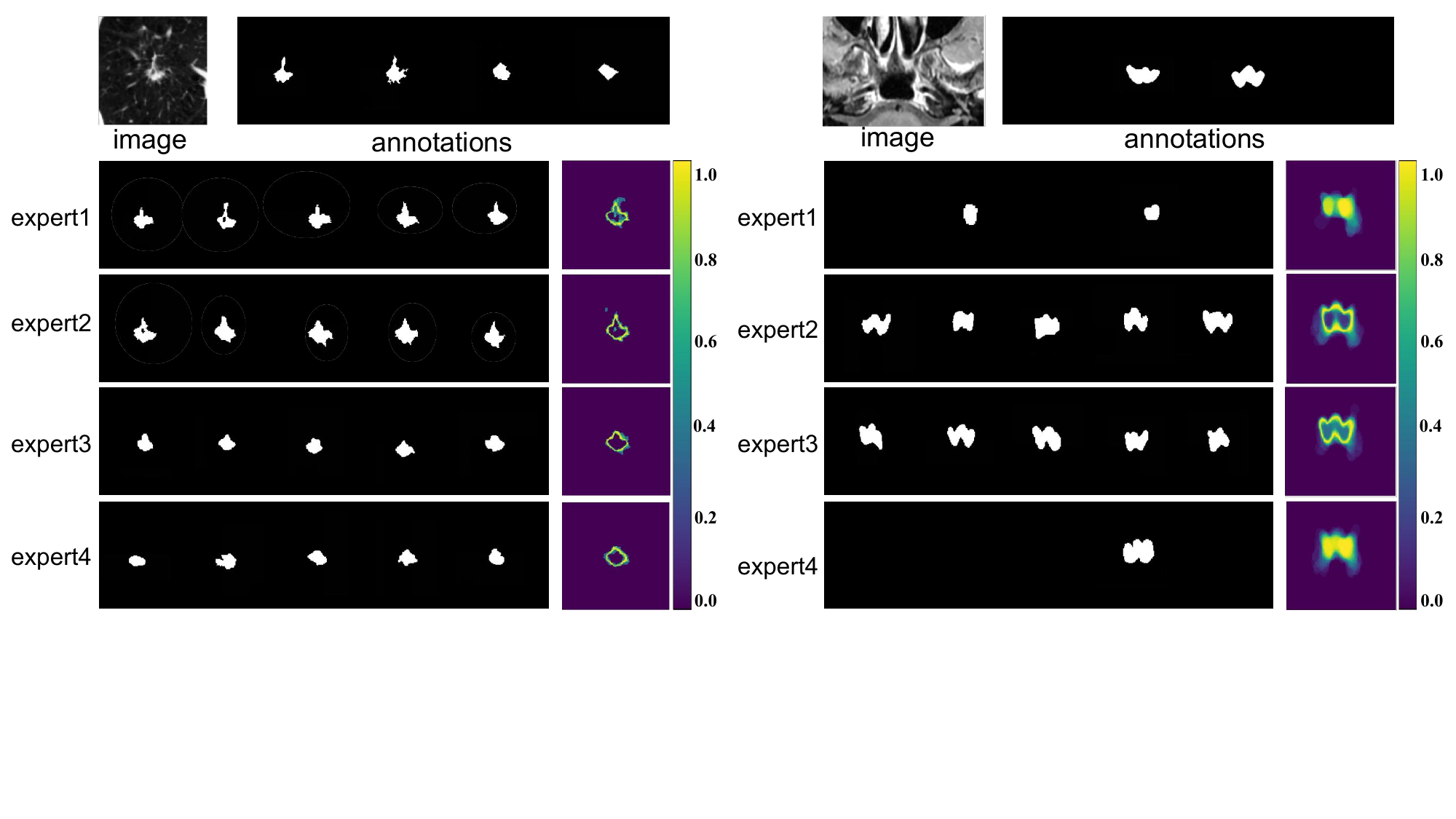}
    \caption{Preference-driven segmentation visualization on both datasets.
    Row 1: original image \& expert annotations (left to right: Experts 1--4). Rows 2-5: Sampled Expert-specific predictions \& uncertainty map (computed by 100 samples).}
    \label{fig:personalized_segmentation}
\end{minipage}
\end{figure}
% \subsubsection{Baselines and Metrics.}
In Stage II, we compare DiffOSeg with CM-Global\cite{ref_cmglobal}, CM-Pixels\cite{ref_cm_pixels}, TAB\cite{ref_TAB}, Pionono\cite{ref_pionono}, and D-Persona. 
To evaluate performance in individualized preference modeling, we quantify segmentation fidelity for each expert using the Dice coefficient ($Dice$, denoted as $D$). This produces $N$+1 metrics: $N$ expert-specific scores ($D_{A_1}, D_{A_2}, \dots, D_{A_N}$) and the overall mean $D_{mean}$ across all experts.

Table~\ref{tab:personality_performance} shows our method achieves consistent improvements. On LIDC-IDRI, DiffOSeg surpasses all baselines for every expert ($A_1-A_4$), demonstrating robust superiority in multi-expert scenarios. On NPC-170, it achieves the best performance for experts ($A_1, A_3$) while marginally underperforming the strongest competitor (D-Persona) on others ($A_2, A_4$). Despite this, DiffOSeg achieves higher $D_{mean}$ than all baselines.
Fig.~\ref{fig:personalized_segmentation} qualitatively demonstrates our model's ability to capture expert-specific styles, showing five randomly sampled predictions and their corresponding uncertainty maps computed from 100 samples.

\subsection{Ablation Study}
\subsubsection{Effect of Probabilistic Consensus Strategy.}
We compare our probabilistic consensus strategy (\textbf{P}) with two alternative approaches: random annotation selection (\textbf{R}) and average annotation aggregation (\textbf{A}). As shown in Table~\ref{tab:ablation1}, our strategy demonstrates superior performance across both $GED$ and $Dice_{soft}$ metrics on LIDC-IDRI. While achieving comparable $GED_{30}$ and $GED_{50}$ to R, our strategy shows clear advantages in segmentation fidelity ($Dice_{soft}$). 
By simulating a wide range of scenarios, the probabilistic consensus
strategy enables robust integration of shared patterns while preserving annotation diversity.
\begin{table}[t]
\centering
\begin{minipage}[t]{0.52\textwidth} % 移除固定高度
\centering
\caption{Consensus-driven segmentation performance of different annotation selection strategies on LIDC-IDRI. R: Random; A: Average; P: Our probabilistic consensus strategy. Best in \textbf{bold}.}
\label{tab:ablation1}
\resizebox{\textwidth}{!}{% % 保持宽度同步
\begin{tabular}{c|cccccc}
\toprule
\textbf{Method} &  $GED_{\scriptscriptstyle 10}(\downarrow)$ & $GED_{\scriptscriptstyle 30}(\downarrow)$  & $GED_{\scriptscriptstyle 50}(\downarrow)$  & $D^{s}_{\scriptscriptstyle 10}(\uparrow)$ & $D^{s}_{\scriptscriptstyle 30}(\uparrow)$ & $D^{s}_{\scriptscriptstyle 50}(\uparrow)$ \\
\midrule
R & 0.0933 & 0.0774 & \textbf{0.0742} & 90.56 & 91.73 & 91.95 \\
A & 0.1194 & 0.1029 & 0.0990 & 89.57 & 90.43 & 90.69 \\
P & \textbf{0.0915} & \textbf{0.0773} & 0.0745 & \textbf{91.21} & \textbf{92.20} & \textbf{92.37} \\
\bottomrule
\end{tabular}%
}
\end{minipage}
\hfill
\begin{minipage}[t]{0.45\textwidth} % 移除固定高度
\centering
\caption{Preference-driven segmentation performance of  different expert identity encoding strategies on LIDC-IDRI. Best in \textbf{bold}.}
\label{tab:ablation2}
\resizebox{\textwidth}{!}{% % 保持宽度同步
\begin{tabular}{c|ccccc}
\toprule
\textbf{Method }
& $D_{A_{1}} (\uparrow)$ & $D_{A_{2}} (\uparrow)$ & $D_{A_{3}} (\uparrow)$ & $D_{A_{4}} (\uparrow)$ & $D_{mean} (\uparrow)$ \\
\midrule
blind  & 90.35 & 88.50 & 89.73 & 89.10 & 89.42 \\
concat. &  90.96 & 89.16 & 90.19 & 89.64 & 89.99 \\
ours &  \textbf{91.41} & \textbf{89.46} & \textbf{90.78} & \textbf{90.09} & \textbf{90.44} \\
\bottomrule
\end{tabular}%
}
\end{minipage}
\end{table}

\subsubsection{Effect of Preference Prompts.}
We systematically evaluate three expert identity encoding strategies: 1) \textbf{blind}: randomly selects annotations from one expert per training step, without incorporating any expert identity information; 2) \textbf{concat.}: incorporates one-hot expert ID vectors via channel concatenation, excluding preference prompts subsequently; 3) \textbf{ours}: our proposed strategy, as described in Sec~\ref{sec:preference_segmentation}.
As evidenced in Table~\ref{tab:ablation2}, our prompt-based approach consistently outperforms baselines across all experts. The learned prompts demonstrate particular effectiveness in capturing nuanced annotation styles, without compromising overall segmentation fidelity.
% \section{Discussion}
% DiffOSeg unifies consensus-driven and preference-driven segmentation through a two-stage diffusion framework, achieving state-of-the-art performance with parameter efficiency (13.1M total vs. D-Persona's 29.4M). 
% Following standard practice in medical multi-rater analysis~\cite{ref_DPersona, ref_cm_pixels, ref_pionono, ref_CIMD}, our method requires multi-expert annotations – clinically aligned yet limiting broader adoption. 
% For future work, we will develop methods to simulate inter-expert variability from single-annotation datasets, potentially expanding clinical applicability.
% \section{Conclusion}
% We propose DiffOSeg, a novel diffusion-based framework that unifies consensus modeling and personalized preference adaptation for multi-rater medical image segmentation. The two-stage architecture first establishes population-level consensus via probabilistic consensus strategy, then adapts preference prompts encoding expert's individual styles.
% Extensive experiments on LIDC-IDRI and NPC-170 demonstrate that DiffOSeg achieves state-of-the-art performance in both tasks. 
\section{Conclusion}
% 添加对标注者风格显式利用的讨论
We propose DiffOSeg, a novel diffusion-based framework that unifies consensus modeling and personalized preference adaptation for multi-rater medical image segmentation. The two-stage architecture first establishes population-level consensus via probabilistic consensus strategy, then adapts preference prompts encoding expert's individual styles.
Extensive experiments on LIDC-IDRI and NPC-170 demonstrate that DiffOSeg achieves state-of-the-art performance in both tasks with parameter efficiency (13.1 M total vs. D-Persona's 28.46 M). Future directions include few-shot adaptation for new unseen experts, and systematic modeling of annotation style variations to enhance clinical diagnostic applications.
\begin{credits}
\subsubsection{\discintname}
The authors have no competing interests to declare that
are relevant to the content of this article.
\end{credits}
% Future directions will include few-shot adaptation for new unseen experts and realistic simulation of multi-rater scenarios from limited annotations.
% For future work, we will develop methods to simulate inter-expert variability from single-annotation datasets, potentially expanding clinical applicability.
\bibliographystyle{splncs04.bst}
\bibliography{Paper-5223.bib}

\begin{thebibliography}{10}
\providecommand{\url}[1]{\texttt{#1}}
\providecommand{\urlprefix}{URL }
\providecommand{\doi}[1]{https://doi.org/#1}

\bibitem{ref_lidc_idri}
Armato~III, S.G., McLennan, G., Bidaut, L., McNitt-Gray, M.F., Meyer, C.R., Reeves, A.P., Zhao, B., Aberle, D.R., Henschke, C.I., Hoffman, E.A., et~al.: The lung image database consortium (lidc) and image database resource initiative (idri): a completed reference database of lung nodules on ct scans. Medical physics  \textbf{38}(2),  915--931 (2011)

\bibitem{ref_phiseg}
Baumgartner, C.F., Tezcan, K.C., Chaitanya, K., H{\"o}tker, A.M., Muehlematter, U.J., Schawkat, K., Becker, A.S., Donati, O., Konukoglu, E.: Phiseg: Capturing uncertainty in medical image segmentation. In: Medical Image Computing and Computer Assisted Intervention--MICCAI 2019: 22nd International Conference, Shenzhen, China, October 13--17, 2019, Proceedings, Part II 22. pp. 119--127. Springer (2019)

\bibitem{ref_lesion_seg}
Carass, A., Roy, S., Jog, A., Cuzzocreo, J.L., Magrath, E., Gherman, A., Button, J., Nguyen, J., Prados, F., Sudre, C.H., et~al.: Longitudinal multiple sclerosis lesion segmentation: resource and challenge. NeuroImage  \textbf{148},  77--102 (2017)

\bibitem{ref_medicalmatting}
Chen, S., Sun, P., Song, Y., Luo, P.: Diffusiondet: Diffusion model for object detection. In: Proceedings of the IEEE/CVF international conference on computer vision. pp. 19830--19843 (2023)

\bibitem{ref_surgeon}
Endo, M., Lin, P.P.: Surgical margins in the management of extremity soft tissue sarcoma. Chinese clinical oncology  \textbf{7}(4),  37--37 (2018)

\bibitem{ref_mose}
Gao, Z., Chen, Y., Zhang, C., He, X.: Modeling multimodal aleatoric uncertainty in segmentation with mixture of stochastic experts. arXiv preprint arXiv:2212.07328  (2022)

\bibitem{ref_ddpm}
Ho, J., Jain, A., Abbeel, P.: Denoising diffusion probabilistic models. Advances in neural information processing systems  \textbf{33},  6840--6851 (2020)

\bibitem{ref_ddim}
Ho, J., Jain, A., Abbeel, P.: Denoising diffusion probabilistic models. Advances in neural information processing systems  \textbf{33},  6840--6851 (2020)

\bibitem{ref_ccdm1}
Hoogeboom, E., Nielsen, D., Jaini, P., Forr{\'e}, P., Welling, M.: Argmax flows and multinomial diffusion: Learning categorical distributions. Advances in neural information processing systems  \textbf{34},  12454--12465 (2021)

\bibitem{ref_vpt}
Jia, M., Tang, L., Chen, B.C., Cardie, C., Belongie, S., Hariharan, B., Lim, S.N.: Visual prompt tuning. In: European conference on computer vision. pp. 709--727. Springer (2022)

\bibitem{ref_prob_unet}
Kohl, S., Romera-Paredes, B., Meyer, C., De~Fauw, J., Ledsam, J.R., Maier-Hein, K., Eslami, S., Jimenez~Rezende, D., Ronneberger, O.: A probabilistic u-net for segmentation of ambiguous images. Advances in neural information processing systems  \textbf{31} (2018)

\bibitem{ref_tabddpm}
Kotelnikov, A., Baranchuk, D., Rubachev, I., Babenko, A.: Tabddpm: Modelling tabular data with diffusion models. In: International Conference on Machine Learning. pp. 17564--17579. PMLR (2023)

\bibitem{ref_fip}
Li, X.L., Liang, P.: Prefix-tuning: Optimizing continuous prompts for generation. arXiv preprint arXiv:2101.00190  (2021)

\bibitem{liao2023transformer}
Liao, Z., Hu, S., Xie, Y., Xia, Y.: Transformer-based annotation bias-aware medical image segmentation. In: International conference on medical image computing and computer-assisted intervention. pp. 24--34. Springer (2023)

\bibitem{ref_TAB}
Liao, Z., Hu, S., Xie, Y., Xia, Y.: Transformer-based annotation bias-aware medical image segmentation. In: International conference on medical image computing and computer-assisted intervention. pp. 24--34. Springer (2023)

\bibitem{liao2024modeling}
Liao, Z., Hu, S., Xie, Y., Xia, Y.: Modeling annotator preference and stochastic annotation error for medical image segmentation. Medical Image Analysis  \textbf{92},  103028 (2024)

\bibitem{ref_cmr}
Luijnenburg, S.E., Robbers-Visser, D., Moelker, A., Vliegen, H.W., Mulder, B.J., Helbing, W.A.: Intra-observer and interobserver variability of biventricular function, volumes and mass in patients with congenital heart disease measured by cmr imaging. The international journal of cardiovascular imaging  \textbf{26},  57--64 (2010)

\bibitem{ref_npc170}
Luo, X., Wang, H., Xu, J., Li, L., Zhao, Y., He, Y., Huang, H., Xiao, J., Song, T., Zhang, S., et~al.: Generalizable magnetic resonance imaging-based nasopharyngeal carcinoma delineation: Bridging gaps across multiple centers and raters with active learning. International Journal of Radiation Oncology* Biology* Physics  (2024)

\bibitem{ref_medsam}
Ma, J., He, Y., Li, F., Han, L., You, C., Wang, B.: Segment anything in medical images. Nature Communications  \textbf{15}(1), ~654 (2024)

\bibitem{ref_promptir}
Potlapalli, V., Zamir, S.W., Khan, S.H., Shahbaz~Khan, F.: Promptir: Prompting for all-in-one image restoration. Advances in Neural Information Processing Systems  \textbf{36},  71275--71293 (2023)

\bibitem{ref_unet}
Ronneberger, O., Fischer, P., Brox, T.: U-net: Convolutional networks for biomedical image segmentation. In: Medical image computing and computer-assisted intervention--MICCAI 2015: 18th international conference, Munich, Germany, October 5-9, 2015, proceedings, part III 18. pp. 234--241. Springer (2015)

\bibitem{ref_pionono}
Schmidt, A., Morales-Alvarez, P., Molina, R.: Probabilistic modeling of inter-and intra-observer variability in medical image segmentation. In: Proceedings of the IEEE/CVF international conference on computer vision. pp. 21097--21106 (2023)

\bibitem{ref_oncologist}
Steenbakkers, R.J., Duppen, J.C., Fitton, I., Deurloo, K.E., Zijp, L., Uitterhoeve, A.L., Rodrigus, P.T., Kramer, G.W., Bussink, J., De~Jaeger, K., et~al.: Observer variation in target volume delineation of lung cancer related to radiation oncologist--computer interaction: a ‘big brother’evaluation. Radiotherapy and Oncology  \textbf{77}(2),  182--190 (2005)

\bibitem{ref_cmglobal}
Tanno, R., Saeedi, A., Sankaranarayanan, S., Alexander, D.C., Silberman, N.: Learning from noisy labels by regularized estimation of annotator confusion. In: Proceedings of the IEEE/CVF conference on computer vision and pattern recognition. pp. 11244--11253 (2019)

\bibitem{ref_staple}
Warfield, S.K., Zou, K.H., Wells, W.M.: Simultaneous truth and performance level estimation (staple): an algorithm for the validation of image segmentation. IEEE transactions on medical imaging  \textbf{23}(7),  903--921 (2004)

\bibitem{ref_DPersona}
Wu, Y., Luo, X., Xu, Z., Guo, X., Ju, L., Ge, Z., Liao, W., Cai, J.: Diversified and personalized multi-rater medical image segmentation. In: Proceedings of the IEEE/CVF Conference on Computer Vision and Pattern Recognition. pp. 11470--11479 (2024)

\bibitem{ref_mmis2024}
Wu, Y., Xie, Y., Luo, X., Wu, Q., Cai, J.: Dataset, challenge, and evaluation for tumor segmentation variability. In: Proceedings of the 32nd ACM International Conference on Multimedia. pp. 11302--11303 (2024)

\bibitem{ref_promptmr}
Xin, B., Ye, M., Axel, L., Metaxas, D.N.: Fill the k-space and refine the image: Prompting for dynamic and multi-contrast mri reconstruction. In: International Workshop on Statistical Atlases and Computational Models of the Heart. pp. 261--273. Springer (2023)

\bibitem{ref_ccdm2}
Zbinden, L., Doorenbos, L., Pissas, T., Huber, A.T., Sznitman, R., M{\'a}rquez-Neila, P.: Stochastic segmentation with conditional categorical diffusion models. In: Proceedings of the IEEE/CVF International Conference on Computer Vision. pp. 1119--1129 (2023)

\bibitem{ref_dodnet}
Zhang, J., Xie, Y., Xia, Y., Shen, C.: Dodnet: Learning to segment multi-organ and tumors from multiple partially labeled datasets. In: Proceedings of the IEEE/CVF conference on computer vision and pattern recognition. pp. 1195--1204 (2021)

\bibitem{ref_cm_pixels}
Zhang, L., Tanno, R., Xu, M.C., Jin, C., Jacob, J., Cicarrelli, O., Barkhof, F., Alexander, D.: Disentangling human error from ground truth in segmentation of medical images. Advances in Neural Information Processing Systems  \textbf{33},  15750--15762 (2020)

\end{thebibliography}

\end{document}